\documentclass[12pt,final]{elsarticle}

\makeatletter
\def\ps@pprintTitle{%
 \let\@oddhead\@empty
 \let\@evenhead\@empty
\def\@oddfoot{\footnotesize
      \itshape\hfill\today}%
     \let\@evenfoot\@oddfoot}

\DeclareGraphicsExtensions{.pdf,.jpeg,.png}
\usepackage[caption = false,font = footnotesize]{subfig}
\usepackage{url}
\usepackage{rotating}

\usepackage[cmex10]{amsmath}
\usepackage{amsfonts}
\usepackage{amssymb}
\usepackage{amsthm}
\usepackage{bm}
\usepackage{empheq}
\usepackage{booktabs}

\usepackage{algorithmic}
\usepackage{setspace}
\usepackage{array}

\usepackage{url}
\usepackage[table]{xcolor}
\usepackage{multirow}
\usepackage{booktabs}
\usepackage{flushend}
\usepackage[colorlinks=true]{hyperref}
\usepackage{float}

\newcommand{\vect}[1]{\mathbf{#1}} 
\newcommand{\norm}[2][2]{\left\lVert #2 \right\rVert_{#1}} 
\DeclareMathSymbol{\R}{\mathalpha}{AMSb}{"52} 

\floatstyle{ruled}
\newfloat{algorithm}{tbp}{loa}
\providecommand{\algorithmname}{Algorithm}
\floatname{algorithm}{\protect\algorithmname}

\definecolor{LightGray}{gray}{0.85}

\journal{Neural Networks}

\begin{document}

\begin{frontmatter}

\title{Adaptation and learning over networks \\ for nonlinear system modeling}

{\footnotesize\begin{center}
       To be published as a chapter in `\textbf{Adaptive Learning Methods for Nonlinear System Modeling}', Elsevier Publishing, Eds. D. Comminiello and J.C. Principe (2018)\end{center}}

\author[sapienza]{Simone Scardapane\corref{cor1}}
\ead{simone.scardapane@uniroma1.it}
\cortext[cor1]{Corresponding author. Phone: +39 06 44585495, Fax: +39 06 4873300.}

\author[xian]{Jie Chen\corref{cor2}}

\author[cotedazur]{C\'edric Richard\corref{cor3}}

\address[sapienza]{Department of Information Engineering, Electronics and Telecommunications, Sapienza University of Rome, Via Eudossiana 18, 00184 Rome, Italy}

\address[xian]{Northwestern Polytechnical University, Xi'an, School of Marine Science and Technology, 127 West Youyi Road, 710072, Xi'an (China)}
\address[cotedazur]{Universit\'e C\^ote d'Azur, Laboratoire Lagrange (UMR CNRS 7293), Parc Valrose, 06108, Nice Cedex 2 (France)}

\begin{abstract}
In this chapter, we analyze nonlinear filtering problems in distributed environments, e.g., sensor networks or peer-to-peer protocols. In these scenarios, the agents in the environment receive measurements in a streaming fashion, and they are required to estimate a common (nonlinear) model by alternating local computations and communications with their neighbors. We focus on the important distinction between single-task problems, where the underlying model is common to all agents, and multitask problems, where each agent might converge to a different model due to, e.g., spatial dependencies or other factors. Currently, most of the literature on distributed learning in the nonlinear case has focused on the single-task case, which may be a strong limitation in real-world scenarios. After introducing the problem and reviewing the existing approaches, we describe a simple kernel-based algorithm tailored for the multitask case. We evaluate the proposal on a simulated benchmark task, and we conclude by detailing currently open problems and lines of research.
\end{abstract}

\begin{keyword}
Nonlinear system identification, distributed systems, adaptive methods, reproducing kernel Hilbert spaces, diffusion algorithms
\end{keyword}

\end{frontmatter}

\section{Introduction}
\label{sec:introduction}

Adaptive filters have been at the heart of digital signal processing over the last century, thanks to their capability of rapidly adapting to streams of incoming data. At the same time, classical filtering approaches have not been satisfactory to handle the challenges posed by large-scale, unstructured big data scenarios that are common today. This has fostered the recent development of techniques to deal with such problems, allowing signal processing to scale to truly massive datasets \cite{cevher2014convex}, and to work with less structured data types, such as graphs \cite{sandryhaila2014big,di2016adaptive}. In this chapter, we look at one peculiar aspect unifying several big data problems, namely, their \textit{distributed} nature, where the data are naturally generated and aggregated at  different locations with possibly poor or expensive network connectivity. Examples of problems in this category abound, including (but not limited to), wireless sensor networks (WSNs) \cite{predd2006distributed}, distributed databases, robotic swarms, fog computing platforms, among others. In all these scenarios, the agents in the network can be severely limited in their capabilities, in terms of either energy constraints (e.g., low-power devices in WSNs), connectivity, privacy, or other aspects. As a consequence, any solution devised for learning and inference over networks needs to be aware of these constraints, making this a challenging problem with wide applications.

Distributed learning can be cast as a decentralized optimization problem, which has a long history in the optimization field \cite{tsitsiklis1986distributed} and in artificial intelligence. In recent years, this problem has gained a renewed interest from the machine learning community, with the development of a number of learning protocols for a variety of models, including boosting \cite{lazarevic2001distributed}, support vector machines \cite{forero2010consensus,scardapane2016distributed}, kernel regression \cite{predd2006distributed}, and sparse linear models \cite{mateos2010distributed,di2013sparse}, to cite a few. Several of these were also applied in signal processing problems, most notably in order to provide distributed inference capabilities in WSNs \cite{predd2006distributed}. A large majority of them, however, is only applicable in \textit{batch} situations, where each agent is allowed to manipulate its entire (local) dataset at each iteration. Distributed filtering algorithms, on the contrary, require the development of online solutions, where the data are received and processed, sequentially, by the agents.

In the filtering literature, a recent series of works was initiated by the development of the so-called `diffusion filtering' (DF) algorithms, starting from the diffusion LMS \cite{lopes2008diffusion} and the diffusion RLS \cite{cattivelli2008diffusion}, up to their more general formulation in terms of generic convex cost functions \cite{chen2012diffusion,sayed2014adaptive,sayed2014adaptation}, which are considered here. DF algorithms are characterized by interleaving local updates in parallel (mimicking classical filters), with communication steps, during which the agents exchange information on their current estimate with their neighbors. Following the development of the main theory, in the subsequent years, several authors have extended classical linear filters to the distributed case (e.g., sparse LMS \cite{di2013sparse} and group LMS \cite{chen2016group}), while others have focused on nonlinear filters, as we review further on in the chapter. For the interested reader, we refer to the guest editorial in \cite{matta2016guest} and references therein for a recent overview of the literature.

\begin{figure}
  \centering
  \includegraphics[scale=1]{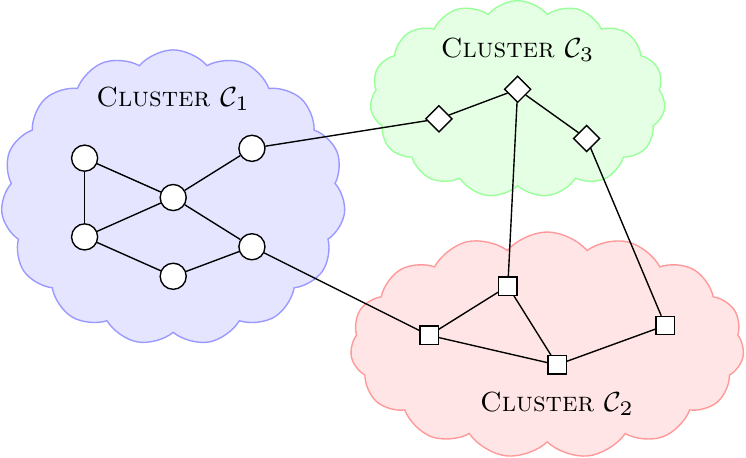}
  \caption{An example of network with $12$ agents subdivided in three different clusters.}
  \label{fig:network_example}
\end{figure}

All the works mentioned up to now assume that the agents are identifying a \textit{common} model. This is a reasonable assumption in several contexts, particularly when the statistics of the data do not depend on the spatial locations of the agents, making it very common in distributed machine learning problems \cite{forero2010consensus,sayed2014adaptation}. In general, however, the agents could be interested in different identification problems, which are similar in some quantifiable sense. As an example, consider the setup illustrated in Fig.~\ref{fig:network_example}. If the agents are sensors deployed over an environment, trying to predict some quantity of interest (e.g., some pollutant concentration), the model might be different among groups (clusters) of agents, possibly due to the spatial conformation of the ground (e.g., sensors deployed over a mountain v.s. sensors deployed in a valley). Nonetheless, since they are all trying to predict the same quantity of interest, communication between agents belonging to different clusters can be beneficial.

The first DF solution to this problem, which is termed \textit{multitask} network, was analyzed in \cite{chen2014multitask}. Following this, a range of algorithms was proposed in the linear case. Most notably, Chen \textit{et al.} \cite{chen2015diffusion} and Zhao and Sayed \cite{zhao2015distributed} investigated the possibility of unsupervised learning of the clusters structure when it is not available \textit{a priori}. Additional developments include the extension to asynchronous networks where, e.g., links may fail or agents might disconnect \cite{nassif2016multitask}; proximal updates for nondifferentiable regularization terms \cite{nassif2016proximal}; total least squares approaches \cite{li2016distributed}; and, finally, multitask learning over (linear) latent subspaces \cite{chen2014diffusion,chen2017multitask}. Almost no work, however, has addressed the problem of learning in a multitask network with \textit{nonlinear} models.

Based on the previous discussion, this chapter has three separate aims. First, we introduce the fundamental concept of DF in Section \ref{sec:problem_formulation}, which serves as a very general introduction to the topic. Next, we summarize recent works on nonlinear DF algorithms in Section \ref{sec:nonlinear_distributed_filtering}, with an emphasis on three classes of solutions. In order to motivate research on multitask learning with nonlinear models, in Section \ref{sec:proposed_algorithm} we propose a multitask kernel algorithm based on a functional formulation of DF. Finally, we validate the algorithm on an experimental benchmark in Section \ref{sec:experimental_evaluation}, before making some final remarks in Section \ref{sec:conclusion}.

\section{Mathematical formulation of the problem}
\label{sec:problem_formulation}

This section is intended to familiarize the reader with some basic theoretical elements underlying most distributed learning scenarios. We start by providing a setup for the problem in Section \ref{sec:problem_setup}. Next, we describe a general class of algorithms based on diffusion protocols in Section \ref{sec:diffusion_algorithms}. In Section \ref{sec:multitask_learning}, we show how these algorithms can be customized to address multitask scenarios. For conciseness, we only focus on a selection of key items, without providing a comprehensive treatment of these thematics. We refer the interested reader to \cite{sayed2014adaptation,sayed2014adaptive} for introductory references.

\subsection{Problem setup}
\label{sec:problem_setup}

Let us consider a generic network of $N$ agents (e.g., sensors in a WSN) as the one depicted in Fig.~\ref{fig:network_example}. We assume that time is slotted and, at every time instant $n$, each agent receives a new observation $\left(\vect{u}_{k,n}, d_{k}(n) \right)$, where $\vect{u}_{k,n} \in \R^M$ is the model input vector at agent $k$ (e.g., a buffer of the last $M$ samples), and $d_{k}(n)$ the corresponding desired response. For simplicity, we assume that $d_k(n)$ is a scalar. For the rest of the chapter, we shall use the subscript $k$ to denote a quantity specific to one of the agents.

Following the standard supervised learning approach, the desired input/output relation can be modeled by choosing a function $f$ in some hypothesis space $\mathcal{H}$. For simplicity, in this section we suppose that each function is parameterized by a vector of tunable parameters $\vect{w} \in \R^q$, e.g., a linear predictor.\footnote{Distributed kernel filters (Section \ref{sec:distributed_kernel_filters}) are an example of a non-parametric formulation, which is recast as a parametric problem thanks to the representer's theorem.} With this setting, each agent is interested in finding a set of parameters $\vect{w}^*_k$ which minimizes some local cost function $J_k(\cdot)$ defined over the hypothesis space from streaming data. Specifically, for most estimation problems in practice, these local cost functions are defined as the expectation of some \textit{error function} $L(\cdot, \cdot)$ with respect to the statistics of the local stream of data:
\begin{equation}
J_k(\vect{w}_k) = \mathbb{E} \biggl\{ L\bigl(d_k, f_k(\vect{u}_k)\bigr) \biggr\} \,.
\end{equation}
where we use the shorthand $f_k(\vect{u}_k) = f(\vect{u}_k; \vect{w}_k)$. The global optimization problem to be solved at the network level is then given by the sum of the local cost functions:
\begin{equation}
\underset{\vect{w}_1, \ldots, \vect{w}_N \in \R^q}{\min} \biggl\{ J_{\text{glob}}(\vect{w}_1, \ldots, \vect{w}_N) \biggr\} = \sum_{k=1}^N J_k(\vect{w}_k) \,,
\label{eq:global_cost_function}
\end{equation}
where $\vect{w}_k$ is the estimate at the $k$th agent. If we assume that no relation holds between the local cost functions, then \eqref{eq:global_cost_function} reduces to a set of $N$ optimization problems that can be solved in parallel by every agent, independently of all the others. A more interesting formulation arises by assuming some form of relation among the cost functions (detailed below). In this case, the information gathered by one agent during its optimization process can potentially be used by the other agents to speedup their convergence, or even converge to a better solution using some shared information.

The difficulty arises from the fact that each agent has direct access to its local cost function, but it has no access to the local cost functions of the other agents for all the reasons mentioned in the introduction. Depending on the relation between cost functions, we can distinguish between three classes of distributed problems:
%
\begin{itemize}
\item{} \textbf{Single-task problems}: in this case, $\vect{w}^*_k = \vect{w}^*, \, k = 1, \ldots, N$, i.e., all the cost functions have the same minimizer which must be attained by all agents. This is the scenario which has drawn most attention in the literature, being particularly useful in distributed machine learning problems \cite{forero2010consensus}, where it is common to assume that the data of interest are generated by a single underlying distribution.
\item{} \textbf{Multitask problems}: in this scenario, each local cost function has possibly a different minimizer $\vect{w}_k$.\footnote{Some readers might recognize that in the machine learning literature, the term `multitask learning' is employed in a slightly different meaning. It refers to the problem of solving several learning tasks defined on the same (or in similar) input domain(s) \cite{evgeniou2004regularized,argyriou2007multi}. While the setup and the objectives in the two cases do not perfectly overlap, we speculate that exploring the connections between them is of particular significance, particularly due to the increasing interest given by deep neural networks \cite{rusu2016progressive}.} In order to make the problem interesting, we assume that these minimizers are `similar' (in some sense to be properly defined) among pairs of neighboring agents, so that communicating can increase their speed of convergence and possibly counter noisy environments.
\item{} \textbf{Clustered multitask problems}: in this intermediate case, each agent belongs to one of $T$ different groups (clusters), such that all agents belonging to the same cluster have the same minimizer, and \textit{vice versa}, as shown in Fig.~\ref{fig:network_example}. Clearly, both single-task and multitask problems can be derived as extreme cases of this class of problems, by setting $T=1$ and $T=N$.
\end{itemize}
%
Multitask problems can be further subdivided, depending on whether the similarity between tasks is known \textit{a priori}, or whether it must be inferred from the data. Examples of the former case are the algorithm in \cite{chen2014multitask}, while examples of the latter case can be found in \cite{chen2015diffusion,zhao2015distributed}. Inferring knowledge about the groups might require the inclusion of some decentralized clustering procedure in the learning process, which is an interesting problem in its own right \cite{jin2006fast}. For simplicity, in this chapter we will focus on the multitask formulation, but we underline that almost all multitask algorithms can be generalized to handle the clustered multitask case, e.g., see \cite{chen2014multitask}. 

\subsection{Diffusion-based algorithms}
\label{sec:diffusion_algorithms}

In order to describe a family of algorithms to solve the previous problem, we first need to define a model of communication between the different agents. Most of the literature focuses on the case where the communication links form a static, undirected, connected graph $\mathcal{G}$. At each time step, the $k$th agent is allowed to communicate with its set of direct neighbors $\mathcal{N}_k$, while it cannot send messages to agents to which it is not directly linked.\footnote{The graph describes all \textit{feasible} communication links. This is a relatively general formulation, since every multi-hop network can be described with an equivalent single-hop network by considering all possible paths as a direct link in the corresponding graph.} Nonetheless, the overall connectedness of the graph ensures that information can flow throughout the entire network. In this scenario, connectivity can be described by a symmetric, real-valued matrix $\vect{A} \in \R^{N \times N}$ such that $A_{kl} \neq 0$ only if agents $k$ and $l$ are connected, and:
\begin{equation}
\sum_{k=1}^N A_{kl} = 1\,, \,\, A_{kl} \ge 0 \,\, \text{ for any } k,l = 1, \ldots, N \,.
\end{equation}
These weights are used by the agents to scale and combine information received by their neighbors. The previous condition ensures that each row of the matrix defines a convex combination, so that the range of the information to be combined is always preserved (formally, the condition requires that $\vect{A}$ be left stochastic). There are several strategies allowing agents to build such matrices, as we show later. This formulation can also be extended in several ways, most notably with the use of asynchronous formulations \cite{zhao2015asynchronous} (thus avoiding the need for a common clock throughout the network), and mixing matrices that do not respect double stochasticity \cite{yuan2017exact}. Nonetheless, since this chapter is only intended as an introduction, we will focus on the simpler case detailed before.

\begin{figure}
  \centering
  \includegraphics[scale=1]{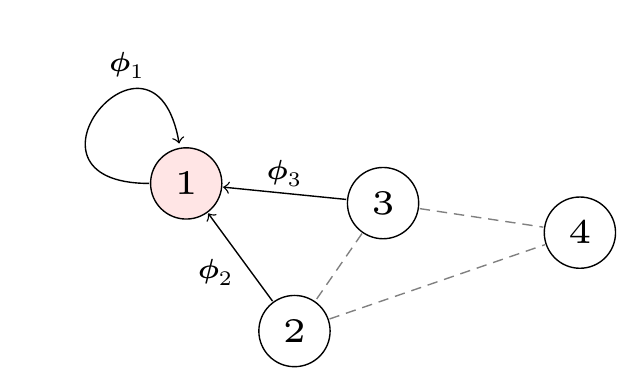}
  \caption{An example of a diffusion step relative to agent $1$ in a simple network with $4$ agents. Gray links are inactive.}
  \label{fig:diffusion_example}
\end{figure}

In the case of a single agent, \eqref{eq:global_cost_function} could be solved by a simple gradient descent algorithm. In order to counteract the lack of global information, the basic idea of diffusion algorithms is to interleave local optimization steps with communication steps, where each agent combines its own estimate with those of its neighbors. In the filtering literature, this strategy is generally denoted as adapt-then-combine (ATC):
\begin{align}
\boldsymbol{\phi}_{k,n} & = \vect{w}_{k,n-1} - \mu_k \nabla J_k(\vect{w}_{k,n-1}) \,, \label{eq:adapt_step}\\
\vect{w}_{k,n} & = \sum_{l=1}^N A_{lk} \boldsymbol{\phi}_{l,n} \label{eq:combine_step}\,,
\end{align}
where $\mu_k$ is a (possibly time-dependent) step size and, similarly to before, we use a double subscript $(k,n)$ to denote the estimate of agent $k$ at time $n$. Practically, the gradient term in \eqref{eq:adapt_step} can be substituted with a noisy version, for example using an instantaneous approximation computed from the current data sample. An example of diffusion step is given in Fig.~\ref{fig:diffusion_example}.

These algorithms are particularly suitable in single-task scenarios, where their convergence properties in the convex case have been analyzed extensively \cite{sayed2014adaptive,sayed2014adaptation}. Interestingly, they can still have good convergence properties in the multitask case, as shown in \cite{chen2015diffusion}. However, they provide a building block for all other formulations, as we show in the next subsection. Before that, we describe an example of diffusion algorithm for nonlinear learning with a distributed logistic regression model.

\subsubsection{An example: logistic regression over networks}

Consider a binary classification problem where $d_{k}(n) \in \left\{0,1\right\}$, i.e., each output is a single bit representing whether the corresponding input belongs to a certain class or not. We approximate the underlying relation using a logistic predictor $f(\vect{u}) = \sigma(\vect{w}^T\vect{u})$, where $\sigma(\cdot)$ is the sigmoid function:
\begin{equation}
\sigma(s) = \frac{1}{1+\exp\left\{-s\right\}} \,,
\end{equation}
ensuring that the outputs of the model are properly scaled as valid probabilities. Each agent wishes to minimize the (regularized) expected cross-entropy over its stream of data:
\begin{equation}
J_k(\vect{w}) = \mathbb{E} \biggl\{ -d_k \cdot \log\bigl(f_k(\vect{u}_k)\bigr) - (1-d_k) \cdot \log\bigl(1-f_k(\vect{u}_k)\bigr) \biggr\} + \frac{\lambda}{2N}\norm{\vect{w}_k}^2 \,,
\end{equation}
where the factor $1/N$ in the regularization term ensures that the total penalization in \eqref{eq:global_cost_function}, when summed over all agents, is equal to $\frac{\lambda}{2}$. By taking instantaneous approximations to the gradient, simple algebra manipulations show that the update steps in \eqref{eq:adapt_step} are given by:
\begin{equation}
\boldsymbol{\phi}_{k,n} = \vect{w}_{k,n-1} + \mu_k \bigl(d_{k}(n) - f_{k,n-1}(\vect{u}_{k,n})\bigr)\vect{u}_{k,n} - \frac{1}{N}\mu_k\lambda\vect{w}_{k,n-1} \,.
\end{equation}
In order to show the speedup obtained by such procedure, in Fig.~\ref{fig:distributed_logistic_regression_example} we plot the average accuracy (see below) obtained with a network of $20$ agents, whose connectivity is generated randomly, and where at every iteration each agent receives a randomly chosen example taken from the well-known Wisconsin Breast Cancer Database (WBCD).\footnote{\url{https://archive.ics.uci.edu/ml/datasets/Breast+Cancer+Wisconsin+(Diagnostic)}} The accuracy is defined as $1$ if the agent makes a correct prediction (i.e., the sign of $f(\vect{u})$ agrees with $d$), $0$ otherwise. It is computed before the adaptation step, and it is averaged with respect to the different agents and the different simulations.

\begin{figure*}
\centering
\subfloat[Network connectivity]{
	\includegraphics[width=0.4\columnwidth,keepaspectratio]{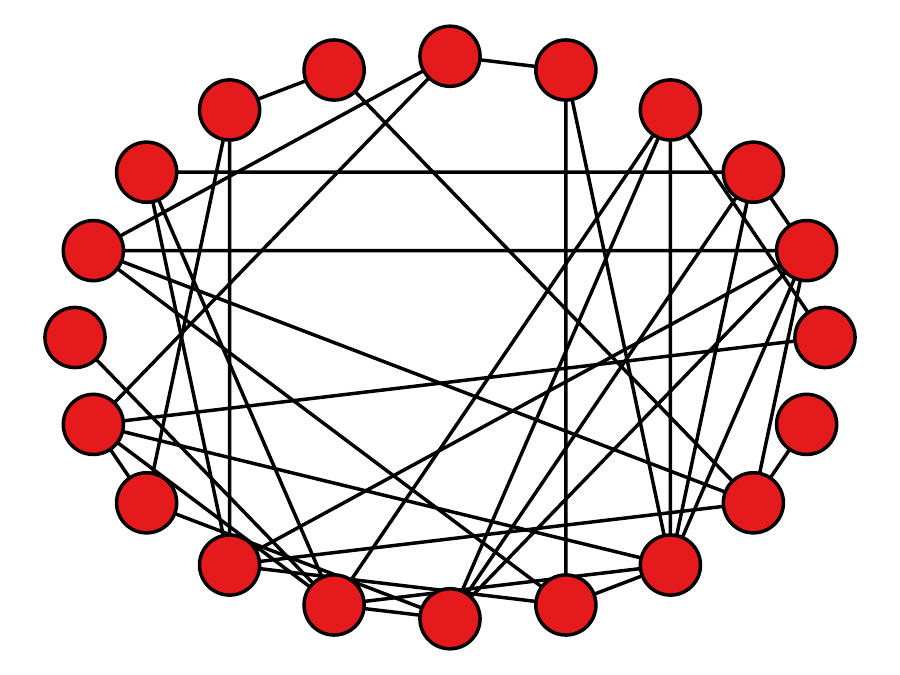} %
} \hfil
\subfloat[Accuracy]{
	\includegraphics[width=0.48\columnwidth,keepaspectratio]{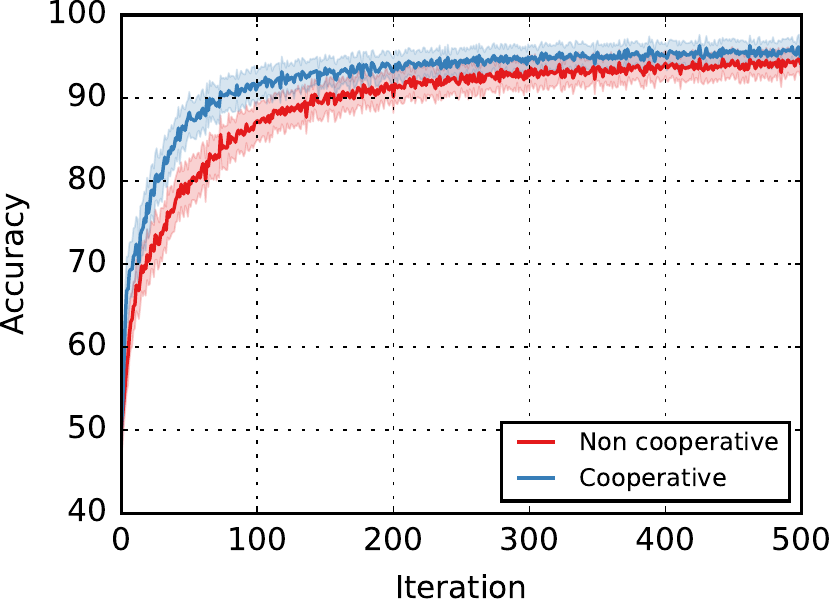} %
} \vfil
{\caption{An example of distributed (online) logistic regression on the WBCD dataset. (a) The network of $20$ agents used in this experiment. (b) Average accuracy for the diffusion algorithm, in blue, and for a network of $N$ non-cooperating agents (which is equivalent to set $\vect{A} = \vect{I}$), in red. We use shaded contours to plot standard deviation.}
\label{fig:distributed_logistic_regression_example}}
\end{figure*}

\subsection{Extension to multitask learning}
\label{sec:multitask_learning}

The general ideas exposed in the previous section can be easily extended in order to be more efficient in the multitask scenario. Here, we detail a simple extension originally proposed in \cite{chen2014multitask} to show one example of such extensions. We refer to the large number of works cited in the introduction for more recent proposals.

Suppose that, given two agents $k$ and $l$, we have a way to quantify the similarity among their respective minimizers. In order to leverage this information, we can augment the original cost function in \eqref{eq:global_cost_function} with a regularization term forcing the similarity among minimizers, in terms of their Euclidean distance:
\begin{equation}
J^{\text{glob}}(\vect{w}_1, \ldots, \vect{w}_N) = \sum_{k=1}^N J_k(\vect{w}_k) 
	+ \eta \sum_{k=1}^N \sum_{l \neq k, l \in \mathcal{N}_k} \rho_{kl} \norm{\vect{w}_k - \vect{w}_l}^2 \,,
\label{eq:global_cost_function_multitask}
\end{equation}
where $\eta$ is a regularization factor, and the nonnegative coefficients $\rho_{kl} \ge 0$ quantify our \textit{a priori} knowledge about the similarities. We assume that, for each agent, the weights are positive and sum to one:
\begin{equation}
\sum_{l=1}^N \rho_{kl} = 1, \,\, \text{ and } \,\, \rho_{kl} = 0 \text{ if } l \notin \mathcal{N}_k, \,\, \forall k \in \left\{1,\ldots,N\right\} \,.
\end{equation}
As a consequence of their definition, these regularization factors mirror the communication network of the agents. Due to this, taking a local optimization step with respect to the estimate of the $k$th agent immediately requires a diffusion step:
\begin{align}
\vect{w}_{k,n} = \vect{w}_{k,n-1}  - \mu_k \nabla J_k(\vect{w}_{k,n-1}) - \mu_k \eta \sum_{l \neq k, l \in \mathcal{N}_k} \frac{\left(\rho_{kl}+\rho_{lk}\right)}{2} \left( \vect{w}_{k,n-1} - \vect{w}_{l,n-1} \right) \,,
\label{eq:multitask_update}
\end{align}
Since the estimates are already exchanged, the previous update step can be implemented without the need for additional combination steps like in the previous section. As an example, by considering a simple linear predictor $f_k(\vect{u}) = \vect{w}_k^T \vect{u}$ and instantaneous approximations to the gradient, we obtain the update rule for the multitask diffusion LMS presented in \cite{chen2014multitask}:
\begin{align}
\vect{w}_{k,n} = \,\, & \vect{w}_{k,n-1}  + \mu_k \bigl( d_k(n) -  \vect{w}_{k,n-1}^T\vect{u}_{k,n} \bigr)\vect{u}_{k,n} - \nonumber \\
& \mu_k \eta \sum_{l \neq k, l \in \mathcal{N}_k} \frac{\left(\rho_{kl}+\rho_{lk}\right)}{2} \left( \vect{w}_{k,n-1} - \vect{w}_{l,n-1} \right) \,.
\end{align}
If we assume that the mixing weights are symmetric, $\frac{\left(\rho_{kl}+\rho_{lk}\right)}{2}$ simplifies to $\rho_{kl}$. It is possible to obtain asymmetric regularization terms by considering a game-theoretical formulation of the optimization problem, see the discussion in \cite{chen2014multitask}.

\section{Existing approaches to nonlinear distributed filtering}
\label{sec:nonlinear_distributed_filtering}

In this section we describe three approaches to extend the previous formulation to nonlinear models in an efficient way. We underline that all these algorithms have been devised mostly for the case of single-task networks. Further extensions to the multitask scenario are the topic of the next section.

\subsection{Expansion over random bases}
\label{sec:expansion_over_random_bases}

One immediate idea is to project the original input vector $\vect{u}$ to a high-dimensional space via some fixed function $\vect{h}(\vect{x}): \R^M \rightarrow \R^B$ before using it with a linear predictor, where $d$ is the dimensionality of the input vector $\vect{u}$, and $B$ is a parameter which (in general) can be chosen by the user. In the distributed case, this requires only a small communication overhead in the beginning for the agents to agree on a specific projection function. Any distributed linear algorithm, such as the diffusion LMS or the diffusion RLS, can then be used.

Generally speaking, deterministic mappings (such as those mentioned in Chapters 2 and 3) are not efficient, because their size might grow exponentially with respect to $M$. A different idea is to use basis functions whose parameters are assigned stochastically, e.g. a parameterized sigmoid:
\[
h_i(\vect{u}) = \frac{1}{1 + \exp\left\{-\vect{a}_i^T\vect{u} - b_i\right\}} \,,
\]
where $\vect{a}_i$ and $b_i$ might be extracted randomly from some uniform distribution (whose range is generally chosen in order to provide a good accuracy, see the discussion in \cite{scardapane2017randomness}). Interestingly, it is possible to show that the resulting estimator (called a random vector functional-link network) is a universal approximator over compact functions provided that $B$ is chosen large enough \cite{igelnik1995stochastic}. It is also possible to interpret it as a degenerate case of the echo state network described in Chapter 12, where connections between different nodes have been removed. The idea of using RVFL networks in a distributed context was proposed in \cite{scardapane2015distributed} for the batch case, and in \cite{huang2015distributed} for the online case with DF algorithms.

A different approach is to design a feature mapping $\vect{h}(\cdot)$ approximating a specific kernel $\mathcal{K}(\cdot, \cdot)$ function\footnote{We refer to Chapters 6-8 for introductory material on kernel filters.} chosen by the user:
\begin{equation}
\mathcal{K}(\vect{u}_1, \vect{u}_2) \approx \langle \vect{h}(\vect{u}_1), \vect{h}(\vect{u}_2) \rangle \,.
\end{equation}
This idea was popularized by \cite{rahimi2007random} for approximating shift-invariant kernels (e.g., the Gaussian kernel) in large-scale applications of kernel methods. In particular, it is possible to show that this class of kernels can be easily approximated with very simple stochastic mappings. \cite{singh2012online} was the first to apply this idea explicitly to kernel filters, and similar algorithms were independently reintroduced in \cite{bouboulis2016efficient}. Since Chapter 8 is entirely devoted to this idea, we will not go further into it. We refer the interested reader to \cite{scardapane2017randomness} for a recent overview on random feature methods.

\subsection{Distributed kernel filters}
\label{sec:distributed_kernel_filters}

An alternative line of research is devoted to distributed strategies for kernel filters, working directly on some reproducing kernel Hilbert space (RKHS), instead of approximating the kernel function as in the previous section. As we stated in the introduction, several distributed algorithms for kernel ridge regression were devised in the context of WSNs \cite{predd2006distributed}, followed by algorithms for the distributed optimization of SVMs \cite{navia2006distributed,forero2010consensus}. Any approach to dealing with kernels faces the challenge of working with a kernel-based model that depends explicitly on all the data in the training set. A na\"ive distributed implementation would thus require to exchange all the local datasets between the agents, which can become infeasible.

In an online context, this is made worse by the growing nature of the kernel model \cite{honeine2008distributed}. This is a fundamental drawback underlying any kernel filter algorithm \cite{richard2009online,parreira2012stochastic,honeine2007line}. An initial investigation in developing a fully distributed version of the kernel LMS (KLMS) was made in \cite{gao2015diffusion}, where the basic idea is to consider diffusion algorithms directly in a functional form. In particular, let us assume that the data received from the $k$th agent satisfies a model of the form:
\begin{equation}
d_{k}(n) = \psi^o_k(\vect{u}_{k,n}) + \nu_{k}(n) \,,
\label{eq:RKHS_data_model}
\end{equation}
where $\psi_k^o$ belongs to a RKHS $\mathcal{H}$, while $\nu_{k}(n)$ is a zero-mean white noise with variance $\sigma_k^2$. Restricting our attention to a generic single-task network, we have:
\[
	\psi_k^o = \psi^o \hspace{0.5em} \forall k \in \left\{ 1, \ldots, N \right\} \,, 
\]
Considering the classical squared error function, the gradient of the local cost functions can now be computed in terms of their Fr\'echet derivatives as:\footnote{A functional derivative is needed because the dimentionality of $\mathcal{H}$ can be infinite. See \cite{bouboulis2011extension} for an introduction to Fr\'echet derivatives in the context of kernel methods, and \cite{balakrishnan2012applications} for an introductory textbook on functional analysis.}
\begin{equation}
\nabla J_k(\psi_k) = -2\mathbb{E}\biggl\{ \bigl( d_k - \psi_k(\vect{u}_k ) \bigr) \kappa(\cdot, \vect{u}_k) \biggr\} \,,
\end{equation}
where $\kappa$ is the kernel function associated to the RKHS. Considering instantaneous approximations for the expectation as was done earlier, we arrive at a functional equivalent of the ATC diffusion framework:
\begin{align}
\delta_{k,n} & = \vect{\psi}_{k,n-1} + \mu_k \left( d_k(n) - \psi_{k,n-1}(\vect{u}_{k,n} ) \right) \kappa(\cdot, \vect{u}_{k,n}) \,, \label{eq:adapt_step_functional}\\
\psi_{k,n} & = \sum_{l=1}^N A_{lk} \delta_{l,n} \label{eq:combine_step_functional}\,,
\end{align}
Although this formulation is extremely general, one has still to solve the problem of the growing structure of the kernel functions. The idea pursued in \cite{gao2015diffusion} is to assume some shared dictionary $\mathcal{D}$ among nodes, whose selection is (at the moment) an open research question. Using this approximation, we can rewrite the desired function as:
\begin{equation}
\psi_{k,n} = \boldsymbol{\beta}_{k,n}^T \vect{k}_{k,n} \,,
\end{equation}
where $\vect{k}_{k,n}$ is the vector of kernel values computed between the current input vector $\vect{u}_{k,n}$ and the shared dictionary $\mathcal{D}$. Each function $\psi_{k,n}$ is now parameterized by the set of linear coefficients $\boldsymbol{\beta}_{k,n}$. The previous algorithm can be rewritten as:\footnote{Note that we consider a simplified formulation with respect to \cite{gao2015diffusion}, where two combination steps are used. Also, we use the same symbol $\boldsymbol{\delta}_k$ for the result of the adaptation step as in \eqref{eq:adapt_step_functional}, but in boldface to underline that it is now a vector-valued quantity.}
\begin{align}
\boldsymbol{\delta}_{k,n} & = \boldsymbol{\beta}_{k,n-1} + \mu_k \left( d_k(n) -  \boldsymbol{\beta}_{k,n-1}^T \vect{k}_{k,n} ) \right) \vect{k}_{k,n} \,, \label{eq:adapt_klms_fixed_dictionary}\\
\boldsymbol{\beta}_{k,n} & = \sum_{l=1}^N A_{lk} \boldsymbol{\delta}_{l,n} \label{eq:combine_klms_fixed_dictionary}\,.
\end{align}
The idea of preselecting a dictionary is not new in the kernel literature. In fact, one of the earliest algorithms for distributed SVMs \cite{navia2006distributed} exploited a similar idea, which is termed semi-parametric SVM. In \cite{chen2014convergence}, a fixed dictionary is used to analyze the convergence behavior of the KLMS algorithm. A derivation of the functional diffusion KLMS algorithm when removing the fixed dictionary constraint is given in \cite{shin2016distributed}. Another extension is presented in \cite{chouvardas2016diffusion}, where a set of consensus constraints is included in the problem to ensure convergence and speedup the algorithm.

\subsection{Diffusion spline filters}
\label{sec:diffusion_spline_filters}

Another possibility for nonlinear learning over networks is given by considering spline adaptive filters (SAFs).\footnote{SAFs are the topic of Chapter 5.} The idea of using SAFs in a distributed environment was recently introduced in \cite{scardapane2016diffusion}. In the following we briefly describe the distributed algorithm. The interested readers can refer to \cite{scarpiniti2013nonlinear,scarpiniti2016steady} for introductory material on the SAF model.

\begin{figure}
\centering
\includegraphics[width=0.6\columnwidth,keepaspectratio]{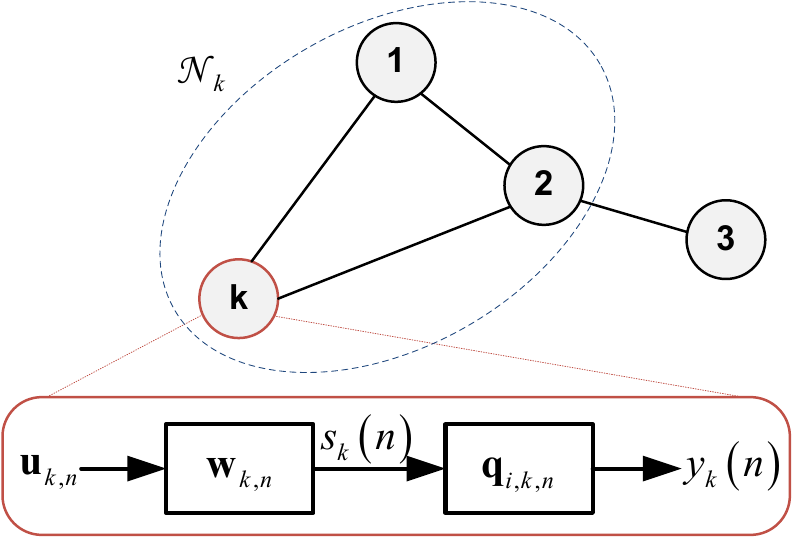}
{\caption{Illustration of SAF interpolation performed over a network of agents (adapted and reprinted with permission from \protect\cite{scardapane2016diffusion}).}
\label{fig:dsaf_network_example}}
\end{figure}

Let us assume that the data are generated according to a restricted Wiener model given by:
\begin{equation}
d_k(n) = f_k^o\left( \vect{w}_k^T \vect{u}_{k,n} \right) + \nu_k(n) \,.
\label{eq:data_model_distributed}
\end{equation}
where $f_k^o$ is any smooth nonlinear function, and $\nu_k(n)$ is a noise term. A SAF mimics this architecture, where the nonlinear term is approximated via spline interpolation over a set of $Q$ fixed control points that are adapted during learning. Hence, in a distributed scenario each agent has estimates of the local part of the filter, $\vect{w}_{k,n}$, and of the aforementioned control points, $\vect{q}_{k,n}$, as shown schematically in Fig.~\ref{fig:dsaf_network_example}.

Consider a single-task scenario, where each agent tries to minimize the expected squared loss. Following \cite{scardapane2016diffusion}, we consider a combine-then-adapt scheme (CTA), where the combination is performed before the adaptation. The two steps are applied to both sets of parameters $\vect{w}_{k,n}$ and $\vect{q}_{k,n}$ simultaneously. However, by exploiting the SAF structure we can avoid exchanging the full weight vector $\vect{q}_{k,n}$, as described in the following. 

The combination step starts with the agents exchanging their current estimates of the linear weights, as:
\begin{align}
\boldsymbol{\psi}_{k,n-1} & = \sum_{l \in \mathcal{N}_{k}} A_{lk} \vect{w}_{l,n-1} \,.
\label{eq:w_diffusion}
\end{align}
The new weights are used to compute the output of the linear part of the filter, denoted as $s_{k}(n) = \boldsymbol{\psi}_{k,n-1}^T \vect{u}_{k,n}$. We use $i$ to denote the index of the closest control point to $s_{k}(n)$ in our set of fixed control points. As described in Chapter 5, the final output of a Wiener SAF depends only on the $i$th control point and its $P$ right neighbors, with $P$ being the order of interpolation. Let us denote by $\vect{q}_{i,k,n-1}$ the set of such `active' control points for agent $k$, which are called the `span' of the filter (see Chapter 5 for more details). We use a third subscript to denote dependence with respect to the span. The second combination step is performed only with respect to the current span:
\begin{align}
\boldsymbol{\xi}_{k,n-1} = \sum_{l \in \mathcal{N}_{k}} A_{lk} \vect{q}_{i,l,n-1} \,.
\label{eq:q_diffusion}
\end{align}
In the case of cubic interpolation, each $\boldsymbol{\xi}_{k,n-1}$ has dimensionality $4$, making its exchange extremely efficient, with only a fixed overhead with respect to a classical diffusion LMS. Practically, every agent sends its current span index $i$ to its neighbors, and receives back the vectors $\vect{q}_{i,l,n-1}$. For simplicity, the mixing weights $A_{lk}$ in the two diffusion steps are assumed identical.

Next, we proceed to the adaptation step. The complete SAF output given the new span is obtained as (again following the general rules of Chapter 5):
\begin{equation}
y_k(n) = \vect{u}^T\vect{B}\boldsymbol{\xi}_{k,n-1} \,.
\end{equation}
where the vector $\vect{u}$ is constructed by taking powers up to a fixed order of the normalized value $\frac{s_k(n)}{\Delta x} - \left\lfloor \frac{s_k(n)}{\Delta x} \right\rfloor$, where $\Delta x$ is the sampling precision of the spline. $\vect{B}$ is the spline matrix, e.g. the Catmull-Rom (CR) spline given by:
\begin{equation}
\vect{B} = \frac{1}{2} 
\begin{bmatrix}
	-1 & 3 & -3 & 1 \\
	2 & -5 & 4 & -1 \\
	-1 & 0 & 1 & 0 \\
	0 & 2 & 0 & 0
\end{bmatrix} \,.
\label{eq:catmulrom_basis_matrix}
\end{equation}
Adaptation is made by performing two parallel gradient descent step:
\begin{align}
\vect{w}_{k,n} & = \boldsymbol{\psi}_{k,n-1} + \mu_k e_{k,n} \varphi'(s_k(n))\vect{u}_{k,n} \,, \\
\vect{q}_{i,k,n} & = \boldsymbol{\xi}_{k,n-1} + \mu_k e_{k,n} \vect{B}^T\vect{u} \,.
\end{align}
where $e_{k,n}$ is the instantaneous local error, and $\varphi'(s_k(n))$ is the spline derivative with respect to the linear weights. Note that the diffusion LMS can be obtained as a special case, where each node initializes its nonlinearity as the identity, and the step size of the nonlinear part is set to zero.

\section{A distributed kernel filter for multitask problems}
\label{sec:proposed_algorithm}

As we saw in the previous section, several ideas have been proposed to model nonlinear systems in a distributed fashion, but almost none is framed for the multi-task scenario. As a first step towards this line of research, in this section we briefly combine some of the previous ideas to devise an efficient kernel-based diffusion algorithm for multi-task networks. In a nutshell, we combine the multi-task diffusion LMS presented in Section \ref{sec:multitask_learning} with the functional diffusion KLMS of Section \ref{sec:distributed_kernel_filters}. To this end, consider again the data model in \eqref{eq:RKHS_data_model}, where we assumed that all the minimizers are the same across the agents. More generally, we can consider the case where two functions $\psi_k^o$ and $\psi_l^o$ are assumed to be `close' in the sense of the norm $\norm[\mathcal{H}]{\cdot}$ of the RKHS, whenever the corresponding agents are spatial neighbors:
\begin{equation}
\psi_k^o \sim \psi_l^o \hspace{0.5em} \text{if } l \in \mathcal{N}_k \,,
\end{equation}
where $\mathcal{N}_k$ denotes the set of neighbors of $k$, and $\sim$ denotes similarity. To recover the unknown functions, and leveraging over the basic idea described in Section \ref{sec:multitask_learning}, we aim at minimizing the following global cost function in a decentralized fashion:
\begin{equation}
J^{\text{glob}}(\psi_1, \ldots, \psi_N) = \sum_{k=1}^N \mathbb{E}\left\{ \bigl\lvert d_{k}(n) - \psi_k(\vect{u}_{k,n}) \bigr\rvert^2 \right\} 
	+ \eta \sum_{k=1}^N \sum_{l \neq k, l \in \mathcal{N}_k} \rho_{kl} \norm[\mathcal{H}]{\psi_k - \psi_l}^2 \,,
\label{eq:J_glob}
\end{equation}
where $\eta > 0$ is a regularization factor, and the nonnegative coefficients $\rho_{kl} \ge 0$ weight the similarity between different functions. Once again, we assume that, for each agent, the weights are positive and sum to one:
\begin{equation}
\sum_{l=1}^N \rho_{kl} = 1, \,\, \text{ and } \,\, \rho_{kl} = 0 \text{ if } l \notin \mathcal{N}_k, \,\, \forall k \in \left\{1,\ldots,N\right\} \,.
\end{equation}
Thus, each agent is interested in minimizing the local expected mean-squared error, under suitable proximity constraints on its function and the functions of its neighbors. The previous problem decomposes as a sum of local cost functions defined as:
\begin{equation}
J^{\text{loc}}_k(\psi_1, \ldots, \psi_N) = \mathbb{E}\left\{ \bigl\lvert d_k(n) - \psi_k(\vect{u}_{k,n}) \bigr\rvert^2 \right\} 
	+ \eta \sum_{l \neq k, l \in \mathcal{N}_k} \rho_{kl} \norm[\mathcal{H}]{\psi_k - \psi_l}^2 \,.
\label{eq:J_loc}
\end{equation}
Each local cost function is independent of the estimate of agents which are not in its immediate neighborhood. Taking the Fr\'echet derivative of \eqref{eq:J_loc} gives us:
\begin{equation}
\nabla J^{\text{loc}}_k(\cdot) = - 2 \mathbb{E}\left\{ \bigl( d_k(n) - \psi_k(\vect{u}_{k,n}) \bigr) \kappa(\cdot, \vect{u}_{k,n}) \right\} + 2\eta \sum_{l \neq k, l \in \mathcal{N}_k} \rho_{kl} \left( \psi_{k,n-1} - \psi_{l,n-1} \right) \,,
\end{equation}
where $\kappa(\cdot, \cdot)$ is the reproducing kernel associated to $\mathcal{H}$. For simplicity, we assume that the mixing weights $\rho_{kl}$ are symmetrical (see the discussion at the end of Section \ref{sec:multitask_learning}). Making an instantaneous approximation for the expectation gives us the following local update rule in functional form at time instant $n$:
\begin{align}
\psi_{k,n} = \psi_{k,n-1} & + \mu_k \left[ d_k(n) - \psi_{k,n-1}(\vect{u}_{k,n})\right]\kappa(\cdot, \vect{u}_{k,n}) \nonumber \\
						  & - \mu_k\eta \sum_{l \neq k, l \in \mathcal{N}_k} \rho_{kl} \left( \psi_{k,n-1} - \psi_{l,n-1} \right) \,,
\label{eq:functional_update}
\end{align}
where the factor $2$ has been included in the step size $\mu_k$. Considering $\mathcal{H}$ as the space of linear predictors over $\vect{u}_{k,n}$, then \eqref{eq:functional_update} reduces to the diffusion LMS for multitask networks presented earlier. In order to have a feasible implementation, once again we assume a shared dictionary $\mathcal{D}$ among agents. \eqref{eq:functional_update} reduces to:
\begin{align}
\boldsymbol{\beta}_{k,n} = \boldsymbol{\beta}_{k,n-1} & + \mu_k \left[ d_k(n) - \boldsymbol{\beta}_{k,n-1}^T\vect{k}_{k,n} \right]\vect{k}_{k,n} \nonumber \\
						  & - \mu_k\eta \sum_{l \neq k, l \in \mathcal{N}_k} \rho_{kl} \left( \boldsymbol{\beta}_{k,n-1} - \boldsymbol{\beta}_{l,n-1} \right) \,.
\end{align}

\section{Experimental evaluation}
\label{sec:experimental_evaluation}

\subsection{Experiment setup}

\begin{figure*}
\centering
\subfloat[Network connectivity]{
	\includegraphics[width=0.45\columnwidth,keepaspectratio]{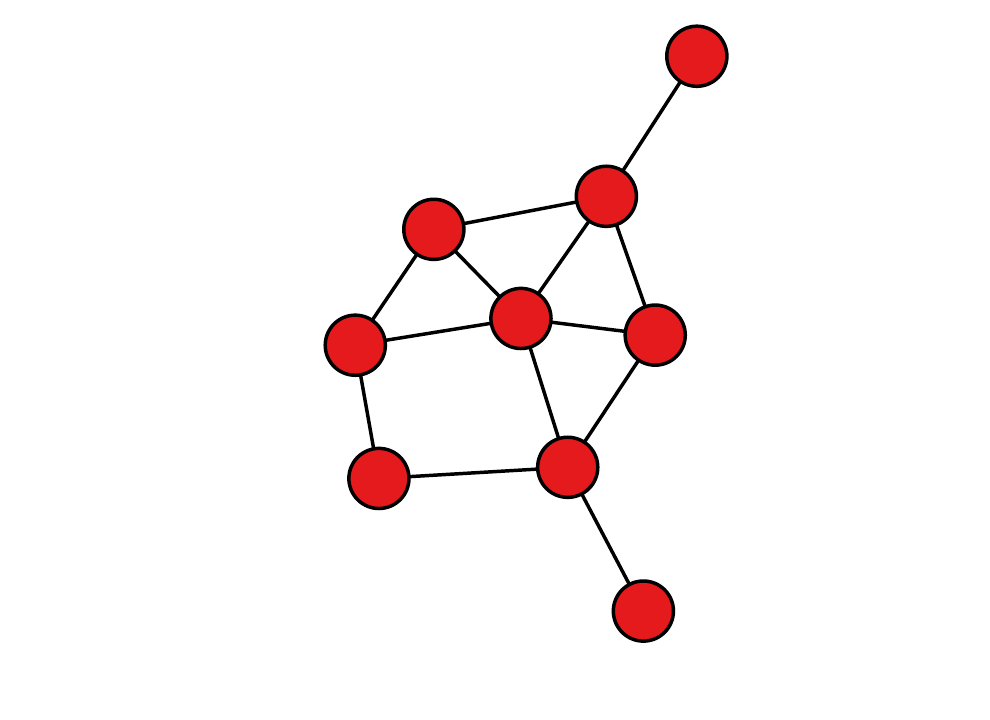} %
} \hfil
\subfloat[Desired output]{
	\includegraphics[width=0.45\columnwidth,keepaspectratio]{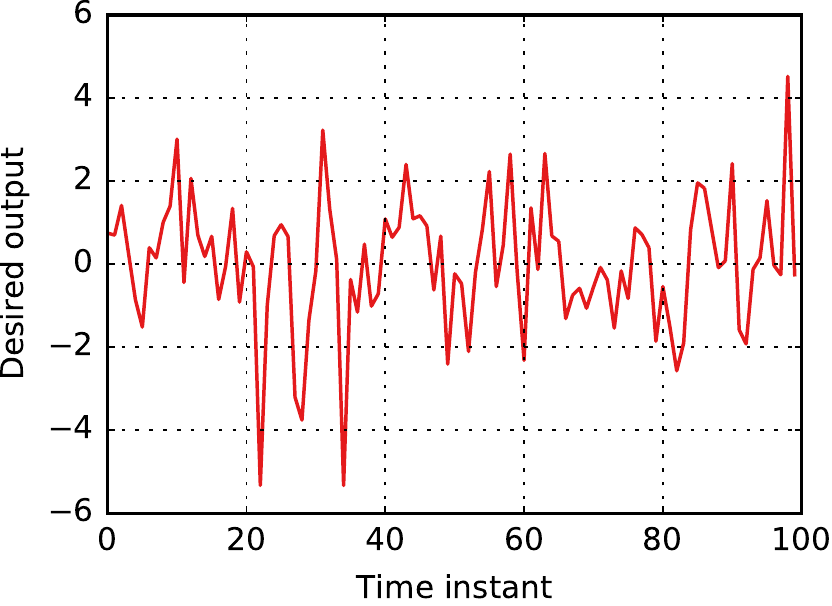} %
} \vfil
\subfloat[Noise variances]{
	\includegraphics[width=0.45\columnwidth,keepaspectratio]{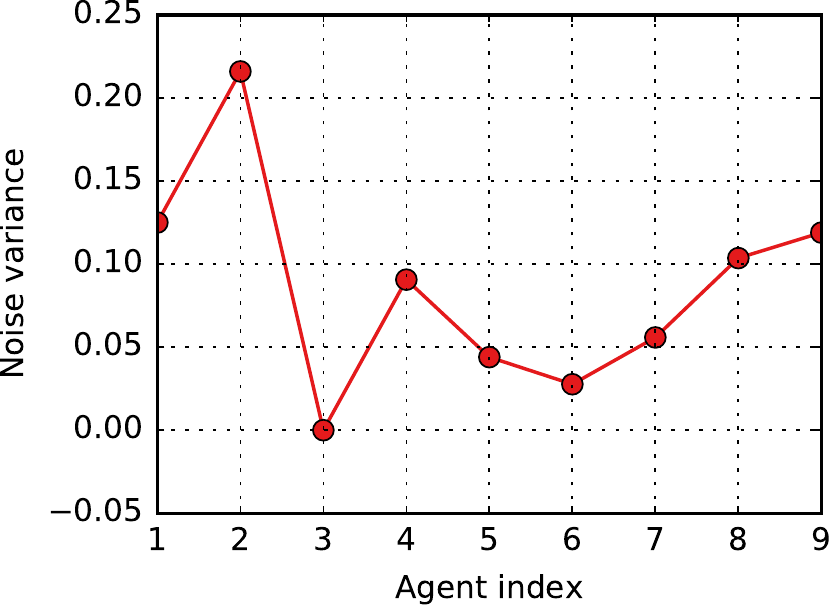} %
} \hfil
\subfloat[Step sizes]{
	\includegraphics[width=0.45\columnwidth,keepaspectratio]{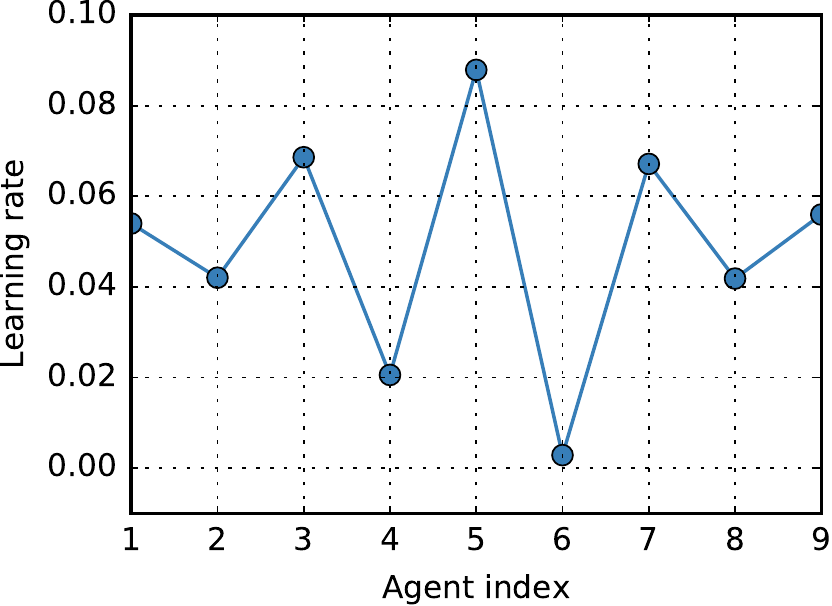} %
} \vfil
{\caption{General setup for the experimental section. (a) The network of $9$ agents used in all experiments. (b) The first $100$ samples of the desired output for the first agent. (c) Noise variance for each agent. (d) Learning rate for each agent. See the text for a description of how (b)-(d) were generated.}
\label{fig:network_setup_experiments}}
\end{figure*}

In this section, we evaluate the proposed method on a simulated multitask nonlinear problem. The output at each agent is given by the following equation:
\begin{equation}
d_{k}(n) = f(\vect{u}_{k,n}) + \vect{w}_{k}^T\vect{u}_{k,n} + \nu_{k}(n) \,,
\label{eq:experiment_data_model}
\end{equation}
which is composed of a common nonlinear part $f(\cdot)$, a local linear part $\vect{w}_{k}^T\vect{u}_{k,n}$, and a local noise of variance $\sigma_k^2$. In particular, we considered a three dimensional input vector $\vect{u} = \left[u_1, u_2, u_3\right]^T$, with the following nonlinearity:
\[
f(\vect{u}) = au_1^2 + bu_2u_3
\]
where $a$ and $b$ were generated from a normal distribution, similarly to the local coefficient vectors $\vect{w}_{k}$. Noise variances were generated uniformly for each agent in the interval $\left[0, 0.3\right]$. We added an additional level of diversity over the network by randomly assigning the learning rates to the agents from the uniform distribution over the interval $\left[0, 0.1\right]$. We considered a network of $9$ agents, whose connectivity was randomly assigned such that each agent is connected in average with one fifth of the other agents, with the requirement that the overall graph is connected. The resulting network connectivity, an example of desired output, and a plot of the noise variances and learning rates, are all shown in Fig.~\ref{fig:network_setup_experiments}. 

We trained the network over a sequence of $1000$ time instants, with white Gaussian inputs with zero mean and unitary variance. The mixing matrix was chosen according to the max-degree heuristic:
\begin{equation}
A_{lk} = 
\begin{cases}
	\frac{1}{\text{deg}_{\max} + 1} & \text{ if $l$ is connected to $k$ } \\
	1 - \frac{\text{deg}_k}{\text{deg}_{\max} + 1} & \text{ if } k = l \\
	0 & \text{ otherwise}
\end{cases}
\label{eq:max_degree_weights} \,,
\end{equation}
where $\text{deg}_k$ is the degree of node $k$, and $\text{deg}_{\max}$ is the maximum degree of the network.\footnote{The degree of a node is the cardinality of the set of its direct neighbors.} Each experiment was averaged over $500$ different runs, by keeping fixed the assignments shown in Fig.~ \ref{fig:network_setup_experiments}.

\subsection{Results and discussion}

\begin{figure}
  \centering
  \includegraphics[scale=0.9]{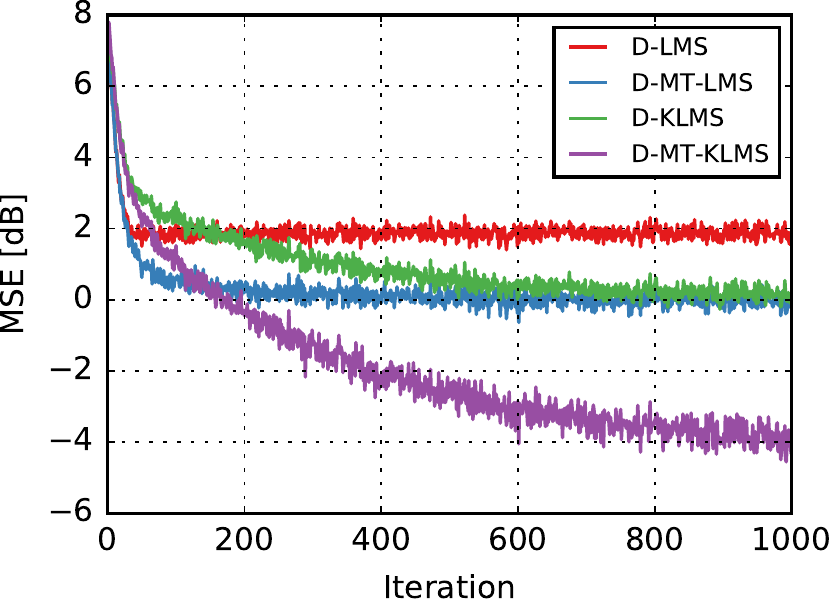}
  {\caption{Average MSE for the algorithms under consideration, averaged both over agents and over $500$ independent runs.}
  \label{fig:accuracy}}
\end{figure}

We compared the performance of a standard diffusion LMS (D-LMS), a multitask D-LMS as described in Section \ref{sec:multitask_learning} (D-MT-LMS), the diffusion KLMS described in Section \ref{sec:distributed_kernel_filters}, and the proposed multitask D-KLMS introduced in Section \ref{sec:proposed_algorithm} (D-MT-KLMS). For the kernel algorithms, we used a Gaussian kernel:
\[
\mathcal{K}(\vect{u}_1, \vect{u}_2) = \exp\Bigl\{ - \gamma \norm{\vect{u}_1 - \vect{u}_2}^2 \Bigr\} \,,
\]
where $\gamma$ was chosen as the inverse of the dimensionality of $\vect{u}$, which was found to provide a good accuracy. For the multitask algorithms, the regularization coefficients were selected uniformly as:
\begin{equation}
\rho_{kl} = 
\begin{cases}
	\frac{1}{\text{deg}_k} & \text{ if $k$ is connected to $l$ } \\
	0 & \text{ otherwise}
\end{cases}
\label{eq:reg_coefficients} \,,
\end{equation}
while the regularization factor was set to $\eta=0.01$. For the kernel algorithms, we fixed \textit{a priori} a dataset of size $100$ with randomly extracted elements. The average MSE in dB across all runs is shown in Fig.~\ref{fig:accuracy}. 

As expected, the D-LMS was the poorest performing algorithm, due to the doubly incorrect assumptions that the agents share the same minimizer, and that the underlying function is linear. By relaxing one of the two assumptions, D-MT-LMS performed better, with an accuracy that is comparable to D-KLMS. Clearly, D-MT-KLMS was the best algorithm in this case, showing that it can be an effective solution for nonlinear multitask problems.

\begin{figure}
  \centering
  \includegraphics[scale=0.9]{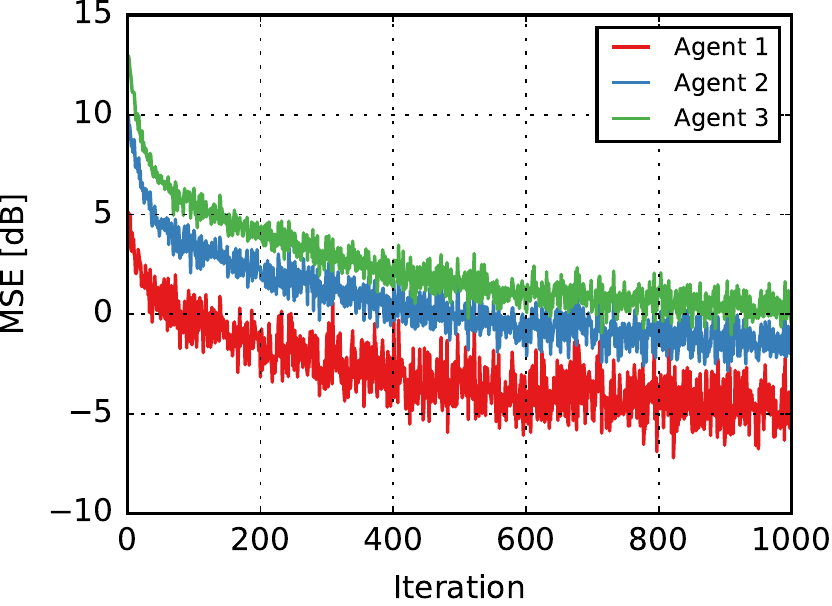}
  {\caption{Local MSE averaged over $100$ runs for $3$ representative agents over the network.}
  \label{fig:mse_local}}
\end{figure}

\begin{figure}
  \centering
  \includegraphics[scale=0.9]{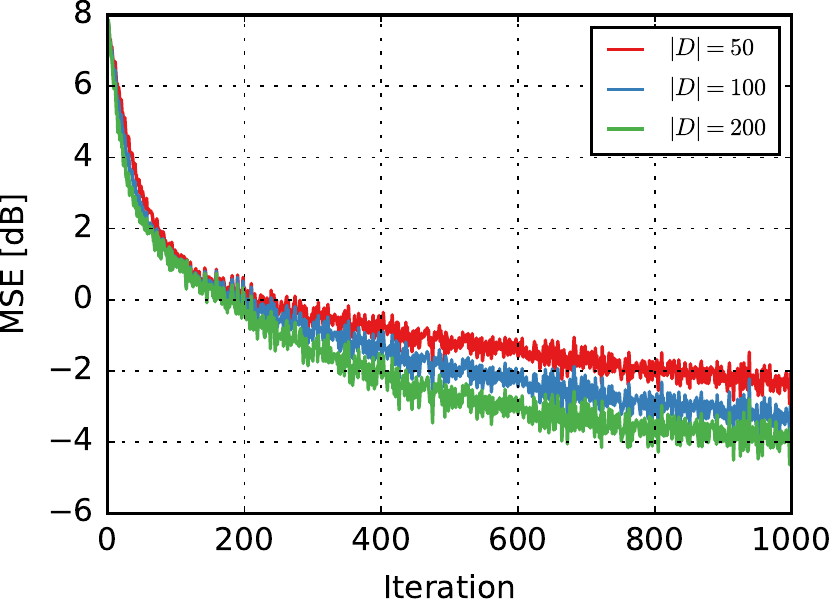}
  {\caption{Average MSE for three different dictionary sizes.}
  \label{fig:accuracy_various_dictionary_sizes}}
\end{figure}

In Fig.~\ref{fig:mse_local} we show the MSE evolution for three representative agents. As expected, their performance is different depending on the selected learning rate and amount of noise, but the multitask algorithm is able to effectively combine the learning curves to obtain the average behavior as in the purple line of Fig.~\ref{fig:accuracy}. Finally, in Fig.~\ref{fig:accuracy_various_dictionary_sizes} we show the average MSE evolution when varying the size of the dictionary. Clearly, increasing the size improves the accuracy (up to a given upper bound), at the cost of a larger computational burden.

\section{Discussion and open problems}
\label{sec:conclusion}

Distributed inference is a fundamental tool according to today's technological trends. In the adaptive filtering community, many classical algorithms can be readily extended to the distributed scenario by exploiting diffusion principles, where local adaptation steps are interleaved with communication steps between neighbors. The resulting algorithms are both computationally efficient, and deployable over a large set of scenarios. In this chapter, we reviewed the basic tools of this field, and we briefly surveyed some of the nonlinear extensions that have been proposed.

An important distinction can be made between single-task problems, where all agents share the same minimizer, and multitask problems, where the minimizers can be different but it is known that they share some similarities. We underlined how very little work has been done on the nonlinear multitask case, and we proposed a simple kernel-based diffusion algorithm to this end. Many extensions over the basic setup of this chapter are possible, most notably a way to remove the assumption of a shared dictionary, an adaptive way to build the regularization coefficients, a theoretical analysis of the algorithm, or additional extensions towards asynchronous networks. Finally, we can consider mixing multitask networks with multi-objective algorithms \cite{chen2013distributed}, such that each agent is interested in minimizing multiple objectives simultaneously.

\section*{Acknowledgments}

The work of Simone Scardapane was supported in part by Italian MIUR, ``\textit{Progetti di Ricerca di Rilevante Interesse Nazionale}'',  GAUChO project, under Grant 2015YPXH4W\_004. The work of Jie Chen was supported in part by the National Natural Science Foundation of China (NSFC grant 61671382).

\bibliographystyle{elsarticle-num}
\bibliography{refs}

\end{document}